\documentclass{article}

\usepackage{arxiv}

\usepackage[utf8]{inputenc} 
\usepackage[T1]{fontenc}    
\usepackage{hyperref}       
\usepackage{url}            
\usepackage{booktabs}       
\usepackage{amsfonts}       
\usepackage{nicefrac}       
\usepackage{microtype}      
\usepackage{lipsum}
\usepackage{graphicx}
\usepackage{subcaption}
\usepackage{algorithm}
\usepackage[noend]{algpseudocode}

\graphicspath{ {./img/} }

\title{Model-Free Episodic Control with Online State Aggregation}

\author{
  Rafael C. Pinto \\
  Instituto Federal de Educação, Ciência e Tecnologia do Rio Grande do Sul\\
  Canoas, RS, Brazil \\
  \texttt{rafael.pinto@canoas.ifrs.edu.br} \\
}

\begin{document}
\maketitle

\begin{abstract}
Episodic control provides a highly sample-efficient method for reinforcement learning while enforcing high memory and computational requirements. This work proposes a simple heuristic for reducing these requirements, and an application to Model-Free Episodic Control (MFEC) is presented. Experiments on Atari games show that this heuristic successfully reduces MFEC computational demands while producing no significant loss of performance when conservative choices of hyperparameters are used. Consequently, episodic control becomes a more feasible option when dealing with reinforcement learning tasks.

\end{abstract}


\section{Introduction}

Reinforcement learning is at the center of many recent accomplishments in artificial intelligence, such as playing Atari games \cite{mnih2015human} and playing \emph{go} at the grandmaster level \cite{silver2016mastering}. Such an approach is very appealing due to its low reliance on supervision, needing only sparse reward signals to acquire useful behaviors. However, most common algorithms suffer from low \emph{sample efficiency}, which means that a high number of training episodes are necessary for the agent to acquire the desired level of competence on diverse tasks. The Deep Q-Learning Network (DQN) \cite{mnih2013playing} and its variants \cite{wang2015dueling,hessel2018rainbow} as well as A3C \cite{mnih2016asynchronous} and other model-free actor-critic or policy gradient methods \cite{schulman2017proximal} may require tens of millions of agent-environment interactions due to very inefficient learning. This kind of inefficiency is not acceptable for some classes of tasks, such as robotics, where failure and damage must be minimized. A robot cannot afford to fall from stairs a thousand times before learning to avoid them. Faster learning is also beneficial for more conventional problems (including games and simulated environments) by allowing for more evaluations and faster iterations. Improvements in sample efficiency are crucial for the advancement of the field and for enabling more practical applications.

It has been hypothesized that gradient-based reinforcement learning methods, such as the above-mentioned ones, suffer from slow learning \cite{botvinick2019reinforcement}. The need for low learning rates prevents the rapid acquisition of new behaviors and immediate incorporation of new information. To offer faster alternatives, non-parametric and instance-based methods have been proposed as replacements \cite{lengyel2008hippocampal, blundell2016model, gershman2017reinforcement, pritzel2017neural}. The main drawback of such approaches is their large memory requirements, as all learning experiences must be stored for later recall. This not only prevents life-long learning but also slows down computations, as large high-dimensional searches are necessary at each step. We propose a way of reducing those requirements by aggregating similar experiences.

This work is structured as follows: section \ref{sec:background} presents related works and concepts in episodic control and state aggregation for reinforcement learning, while section \ref{sec:samfec} presents our proposed algorithm, which introduces state aggregation into episodic control. Section \ref{sec:experiments} shows experimental results, and section \ref{sec:conclusions} concludes with discussions and future works.

\section{Background}
\label{sec:background}

The present work is built upon the concepts of episodic control and state aggregation in reinforcement learning, which shall be reviewed in the next subsections.

\subsection{Episodic Control}

In \cite{lengyel2008hippocampal}, the case for episodic control is made. Its differences to other kinds of control as well as its advantages and disadvantages are described and theoretically analyzed. It is shown that episodic control bests model-based (noisy) learning in various circumstances in the low data regime.

Model-Free Episodic Control (MFEC) \cite{blundell2016model} was proposed as a highly sample-efficient alternative to deep reinforcement learning \cite{mnih2013playing} by storing \emph{all} unique observations encountered by the agent, each one associated with its predicted Q-value and stored in their respective action buffers (there is one buffer per possible action). Action selection is performed by finding the $k$ Nearest Neighbors (kNN) from the current observation in each action buffer. By averaging the Q-values of the $k$ observations in each buffer, it is possible to select the action with the largest predicted value.

Neural Episodic Control (NEC) \cite{pritzel2017neural} replaces vanilla kNN with distance weighted kNN, resulting in a differentiable module that can be integrated into neural networks. As a consequence, convolutional layers can be used to extract better state embeddings to be passed as inputs to episodic control (while MFEC uses random projections \cite{johnson1984extensions, bingham2001random} or Variational Autoencoders \cite{kingma2013auto} for this purpose). Our methods could be applied to NEC as well, but we focus on MFEC in this work.

Finally, \cite{gershman2017reinforcement} provides an extensive review of episodic control and its comparison to model-free and model-based learning, including parallels in neuroscience.

\subsection{State Aggregation}

State aggregation was proposed as a way to reduce computational demands in reinforcement learning problems with continuous state spaces \cite{singh1995reinforcement}. The essential idea consists in grouping states which should be treated in the same way by the algorithm, i.e., similar states with similar Q-values / policy outputs. As an example, the Growing Neural Gas (GNG) algorithm \cite{fritzke1995growing} has been applied as a state quantizer in reinforcement learning \cite{baumann2011state, baumann2012improved}, aggregating and splitting states online as necessary. It is important to note that the authors acknowledge the necessity of considering both input (state) and output (Q-values) spaces when merging states, something that is also part of our proposal. Very similar states which require different actions should stay separated and updated differently, otherwise, a phenomenon known as \emph{perceptual aliasing} \cite{chrisman1992reinforcement} occurs (in this case, in the aggregated state space).

In \cite{agostinelli2019memory}, the authors propose a memory-efficient variant of MFEC where the least recently used (LRU) policy used for replacing observations when a buffer is full is replaced by online clustering (which is a form of state aggregation). Our proposal differs from this one in the sense that aggregation occurs early during learning, as soon as new observations arrive, and not only when a buffer is full.

\section{MFEC with State Aggregation}
\label{sec:samfec}

Here we propose a simple heuristic for merging states as soon as they are observed. Given two new non-negative real-valued hyper-parameters $\varepsilon_{in}$ and $\varepsilon_{out}$, a new observed state-action entry is added to its respective buffer if, and only if, the Euclidean distance from it to its nearest stored neighbor in that buffer is larger than the threshold $\varepsilon_{in}$, or the absolute difference between its Q-value estimate and the one stored in its nearest neighbor entry is larger than $\varepsilon_{out}$. By setting $\varepsilon_{in}$ to $0$, we get vanilla MFEC (all unique states are added). Smaller values of $\varepsilon_{in}$ and $\varepsilon_{out}$ result in higher memory requirements (and thus higher computational demand, due to kNN lookup), while larger values produce more aggressive state aggregation, resulting in smaller buffers and faster lookups. A detailed description of this procedure is shown in algorithm \ref{alg:samfec}.

\begin{algorithm}[htb]
\caption{Model-free episodic control with state aggregation.}
\hspace*{\algorithmicindent} \textbf{Input:} $\varepsilon$ (exploration rate), $k$ (number of neighbors), $\varepsilon_{in}$ (input threshold), $\varepsilon_{out}$ (output threshold)
\begin{algorithmic}[1]
\For{each episode}
    \For{$t = 1, 2, 3, . . . , T$} \Comment{Run episode}
        \State Receive observation $o_t$ from environment.
        \State Let $s_t = \phi(o_t)$. \Comment{Random projection or other kind of embedding}
        
        \If {$X \sim \mathcal{U}(0,1) < \varepsilon$}   \Comment{$\varepsilon$-greedy policy}
            \State Select random action $a_t$ uniformly from the set of all available actions
        \Else   \Comment{Select greedy action}
            \For {each action $a$}
                \If {$(s_t,a) \in Q^{EC}$}  \Comment{If an exact match is found}
                    \State $\widehat{Q^{EC}}(s_t, a) \leftarrow Q^{EC}(s_t,a)$  \Comment{Use single entry estimate}
                \Else   \Comment{Otherwise, compute kNN estimate}
                    \State $\widehat{Q^{EC}}(s_t, a) \leftarrow 0, sum \leftarrow 0$
                    \For{$i = 1, 2, ..., k$ states $s^{(i)}$ nearest to $s_t$}
                        \State $\widehat{Q^{EC}}(s_t, a) \leftarrow \widehat{Q^{EC}}(s_t, a) + Q^{EC}(s^{(i)},a) \times C^{EC}(s^{(i)},a)$ \Comment{Aggregated kNN weighting}
                        \State $sum \leftarrow sum +  C^{EC}(s,a)$
                        \If{$sum >= k$} \Comment{Aggregated kNN radius truncation}
                            \State break
                        \EndIf
                    \EndFor
                    \State $\widehat{Q^{EC}}(s_t, a) \leftarrow \frac{\widehat{Q^{EC}}(s_t, a)}{sum} $ \Comment{Corrected kNN estimate}
                \EndIf
            \EndFor
            \State Let $a_t = arg max_a \widehat{Q^{EC}}(s_t, a)$ \Comment{Greedy policy (could be $\varepsilon$-greedy)}
        \EndIf
        \State Take action $a_t$, receive reward $r_{t+1}$
    \EndFor
    \For{$t = T, T - 1, . . . , 1$} \Comment{For each stored experience, backwards (at end of episode)}
        \State $s_{nearest} \leftarrow argmin_s \|s - s_t\|, \forall s \in $ buffer for action $a_t$  \Comment{Find nearest}
        \If {$s_t = s_{nearest}$} \Comment{If exact match}
            \State $Q^{EC}(s_{nearest}, a_t) \leftarrow \max\{{ Q^{EC}(s_{nearest}, a_t), R_t}\}$  \Comment{Replace Q}
        \ElsIf {$\|s_t - s_{nearest}\| < \varepsilon_{in}$ and $\|R_t - Q^{EC}(s_{nearest}, a_t)\| < \varepsilon_{out}$}  \Comment{If close enough}
                \State $C^{EC}(s_{nearest}, a_t) \leftarrow C^{EC}(s_{nearest}, a_t) + 1$ \Comment{Increment counter}
                \State $\eta \leftarrow \frac{1}{C^{EC}(s_{nearest}, a_t)}$  \Comment{Compute learning rate}
                \State $Q^{EC}(s_{nearest}, a_t) \leftarrow  Q^{EC}(s_{nearest}, a_t) + \eta \left( R_t - Q^{EC}(s_{nearest}, a_t) \right)$  \Comment{Update Q}
                \State $s_{nearest} \leftarrow s_{nearest} + \eta \left( s_t - s_{nearest} \right)$  \Comment{Update state}
        \Else   \Comment{New state is far away}
            \State $Q^{EC}(s_t, a_t) \leftarrow R_t$  \Comment{Create new entry}
            \State $C^{EC}(s_{nearest}, a_t) \leftarrow 1$ \Comment{Initialize counter}
        \EndIf
    \EndFor
\EndFor
\end{algorithmic}
\label{alg:samfec}
\end{algorithm}

\begin{figure}[htb]
    \centering
    \begin{subfigure}{.3\textwidth}
        \centering
        \includegraphics[scale=0.3]{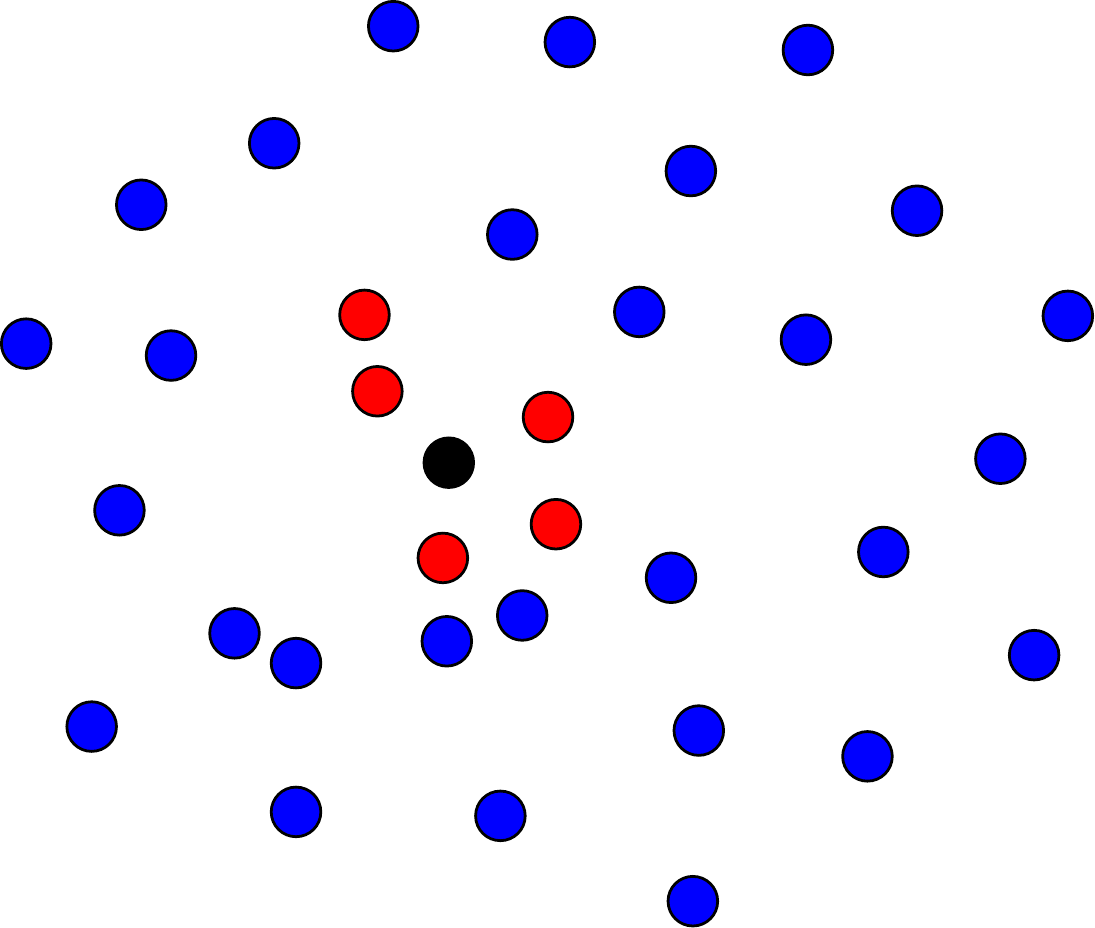}
        \caption{Vanilla kNN.}
        \label{fig:knn1}
    \end{subfigure}
    \begin{subfigure}{.3\textwidth}
        \centering
        \includegraphics[scale=0.3]{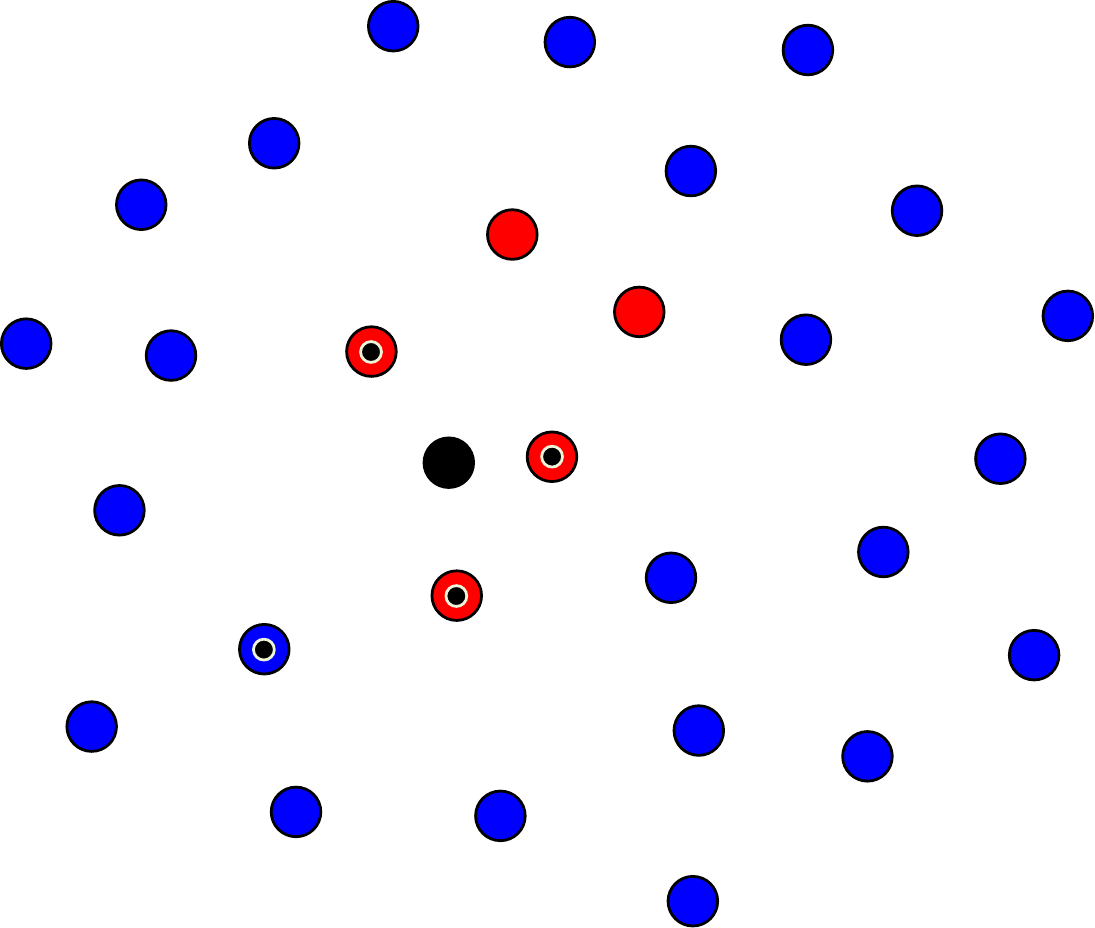}
        \caption{kNN with state aggregation.}
        \label{fig:knn2}
    \end{subfigure}
    \begin{subfigure}{.3\textwidth}
        \centering
        \includegraphics[scale=0.3]{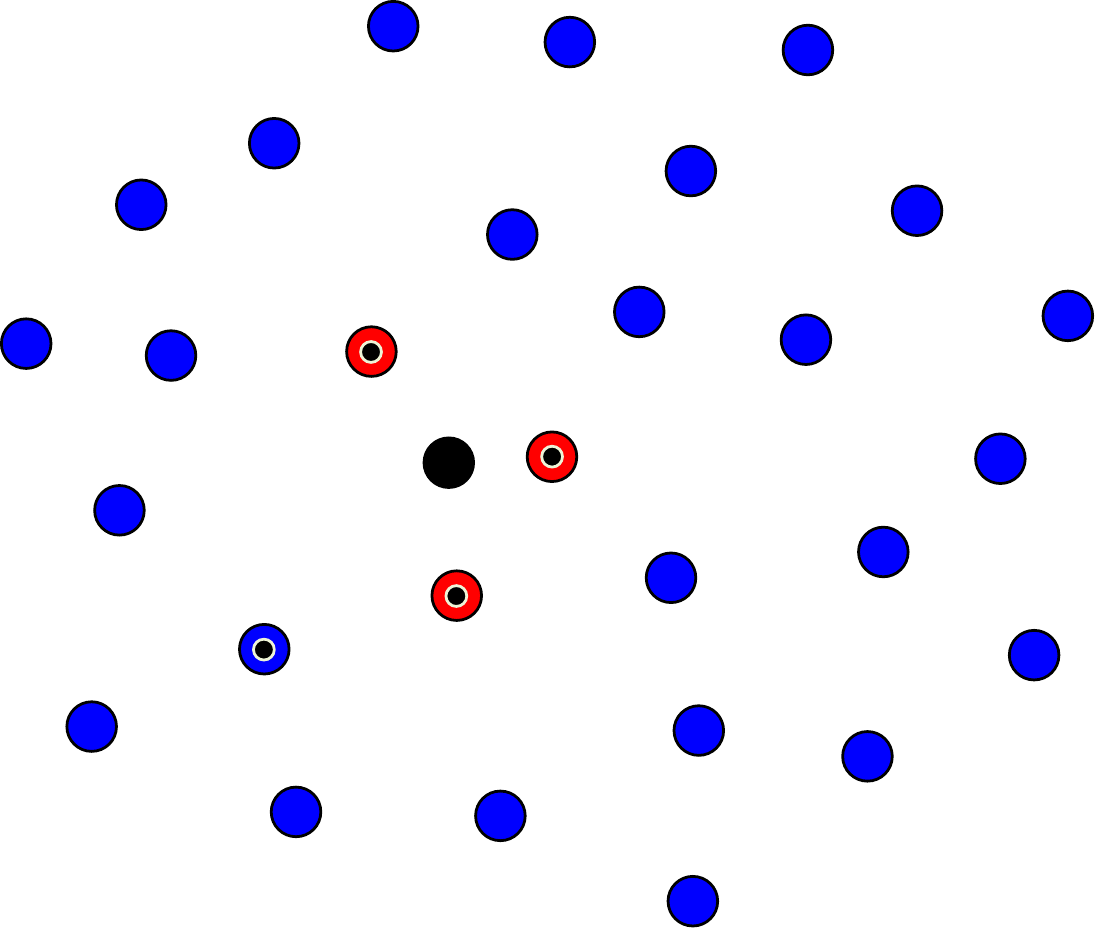}
        \caption{$k$ correction.}
        \label{fig:knn3}
    \end{subfigure}
    \caption{kNN behaves differently with state aggregation for the same $k$. a) kNN ($k=5$) without state aggregation. Red dots represent the $k$ selected points closest to the query point (in black). b) kNN ($k=5$) with state aggregation. Aggregated states are marked with an inner black dot. Note that, in this case, the effective radius considered is increased and, as each point still has the same weight, the center of mass is also distorted, producing very discrepant results when computing the kNN average. c) Our proposed solution: aggregated states count as multiple points. Since the total count of the three nearest states (i.e., $7$) already surpasses $k=5$, no other points are considered. A weighted average is computed by weighting each point according to its count. The effective radius and center of mass are kept similar to vanilla kNN.}
    \label{fig:knn}
\end{figure}

We also note that to preserve the original behavior of kNN, aggregated states should count multiple times when selecting the nearest neighbors. For instance, three aggregated states with counts (number of individual states aggregated into a single entry) $2$, $2$ and $3$, respectively, contain a total of $7$ individual states and thus should be enough when searching for any $k \leq 7$ neighbors, while the first two are enough for $k \leq 4$. Besides minimizing neighborhood distortions in relation to vanilla kNN, correct weights are attributed to each aggregated state, minimizing differences in the final average. Figure \ref{fig:knn} illustrates the problem and the proposed solution.

\section{Experiments}
\label{sec:experiments}

In this section, we describe our experimental setup and results.

\subsection{Setup}
To investigate the pros and cons of performing state aggregation in episodic control, we applied our algorithm to 6 Atari games with 4 different $\varepsilon_{in}$ values, including $0$, which equates to vanilla MFEC. $\varepsilon_{out}$ was set to $100$, $k$ was set to $11$ and $\varepsilon$ (exploration rate) was set to $0.005$ for all experiments (no parameter search was performed). A random projection was performed from a downscaled ($84$x$84$, bilinear) grayscale transformation of the game screen into a $128$-dimensional input vector. Average and peak performance, as well as total buffer size (the sum of all action buffers), were assessed every 100000 frames or so (episodes ran until finished even if the limit was reached). Each game ran for 100 assessments (10 million frames) and this was repeated for 5 sequential seeds from the set ${0,1,2,3,4}$. We are aware of works such as \cite{henderson2018deep} which show that more evaluations are needed to obtain more meaningful results and to avoid effects of handpicking seeds, but due to hardware and time restrictions, we argue that using 5 sequential seeds such as those is enough to avoid the handpicking issue. The "v0" version of each game was used, with no forms of non-determinism (as is the default in previous works on episodic control) and with frameskip set to $4$ (no frame stacking was used, thus $75\%$ of all frames are lost). Our code was implemented in PyTorch \cite{paszke2019pytorch} and will soon be available on GitHub \footnote{Under user  \url{https://github.com/rafaelcp/}}. 

\begin{figure}[htb]
    \centering
    \begin{subfigure}{.49\textwidth}
        \centering
        \includegraphics[width=1\textwidth]{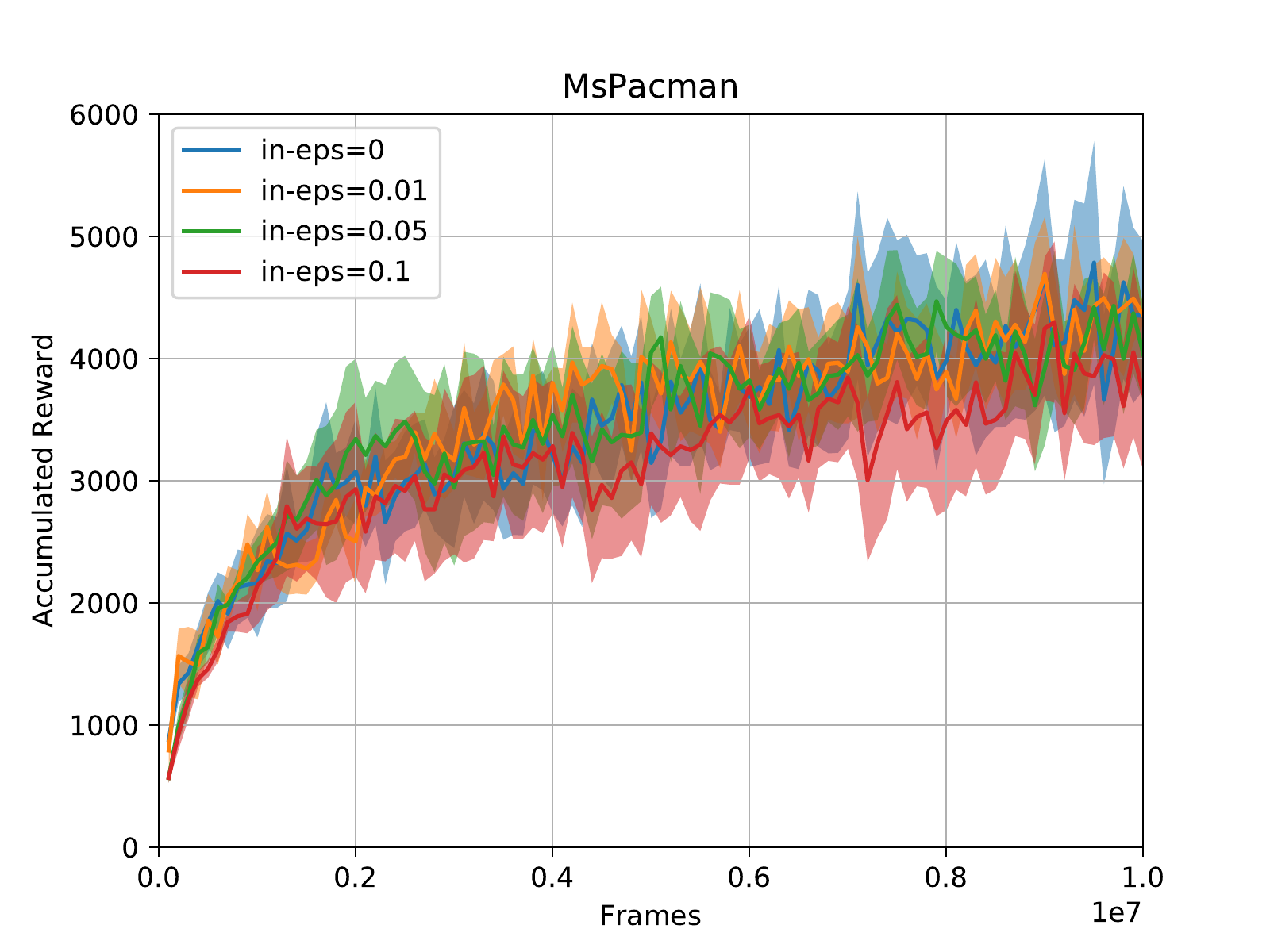}
        \label{fig:perf-mspacman}
    \end{subfigure}
    \begin{subfigure}{.49\textwidth}
        \centering
        \includegraphics[width=1\textwidth]{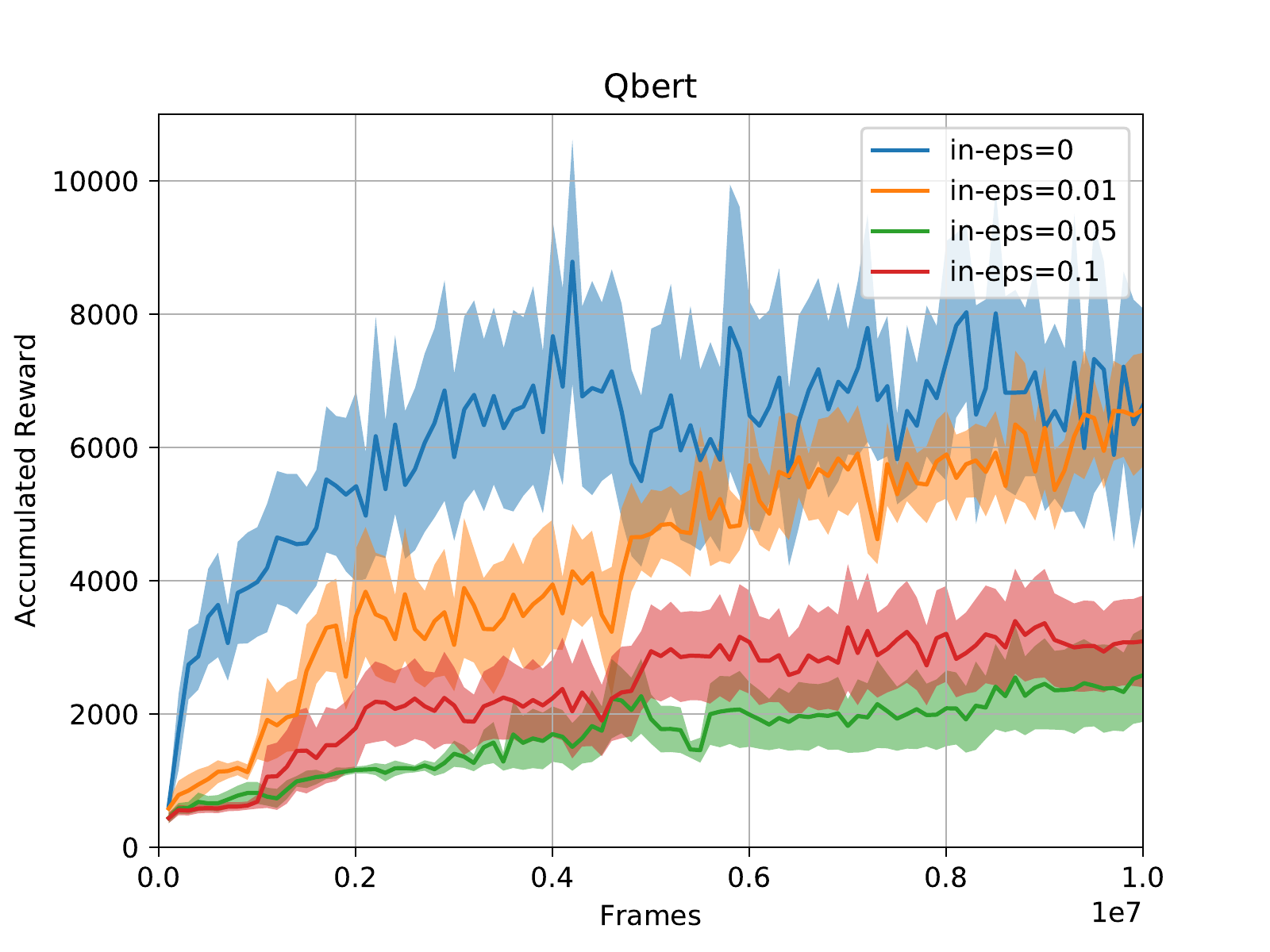}
        \label{fig:perf-qbert}
    \end{subfigure}\\[-6.4ex]
    \begin{subfigure}{.49\textwidth}
        \centering
        \includegraphics[width=1\textwidth]{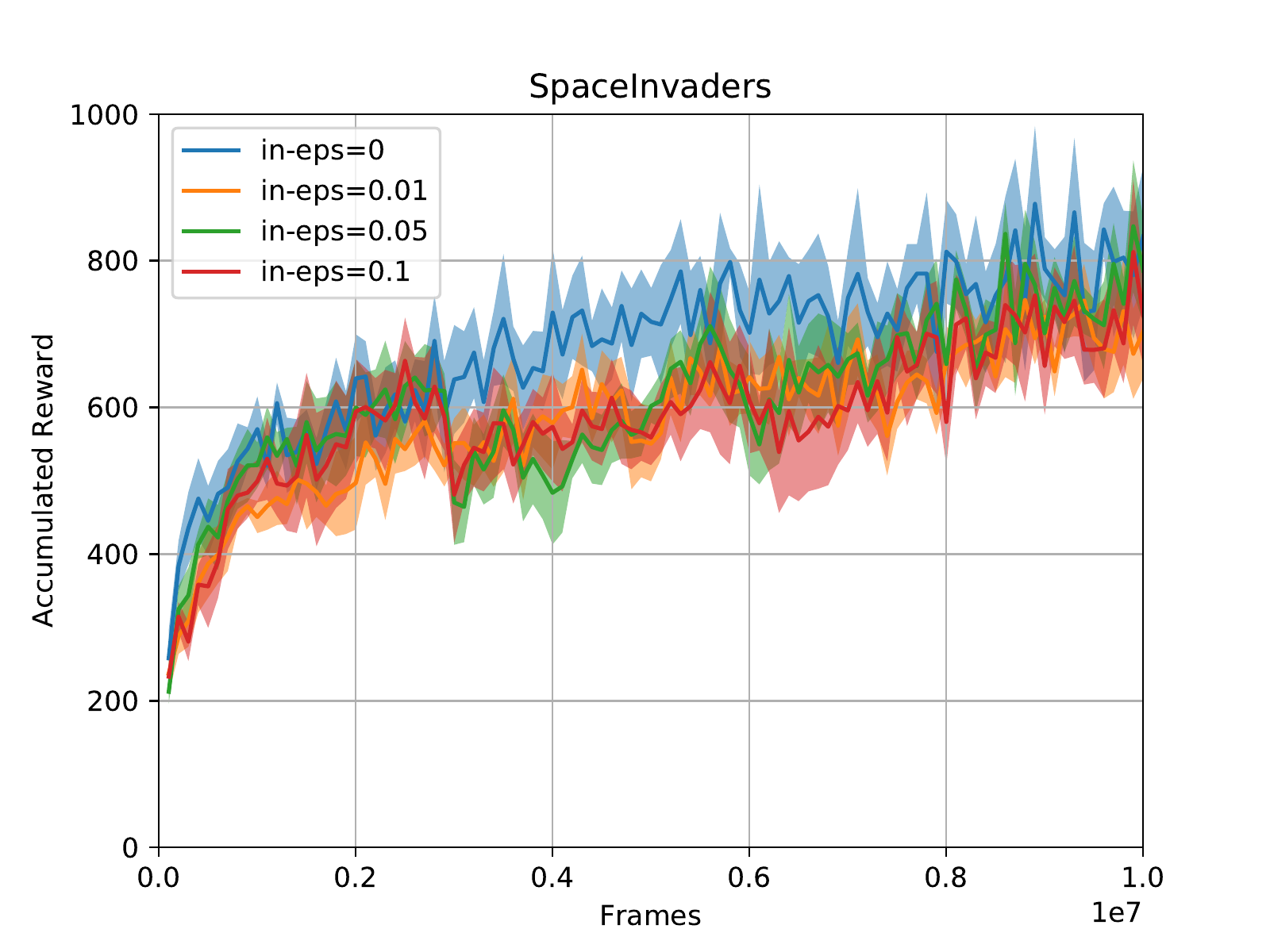}
        \label{fig:perf-spaceinvaders}
    \end{subfigure}
    \begin{subfigure}{.49\textwidth}
        \centering
        \includegraphics[width=1\textwidth]{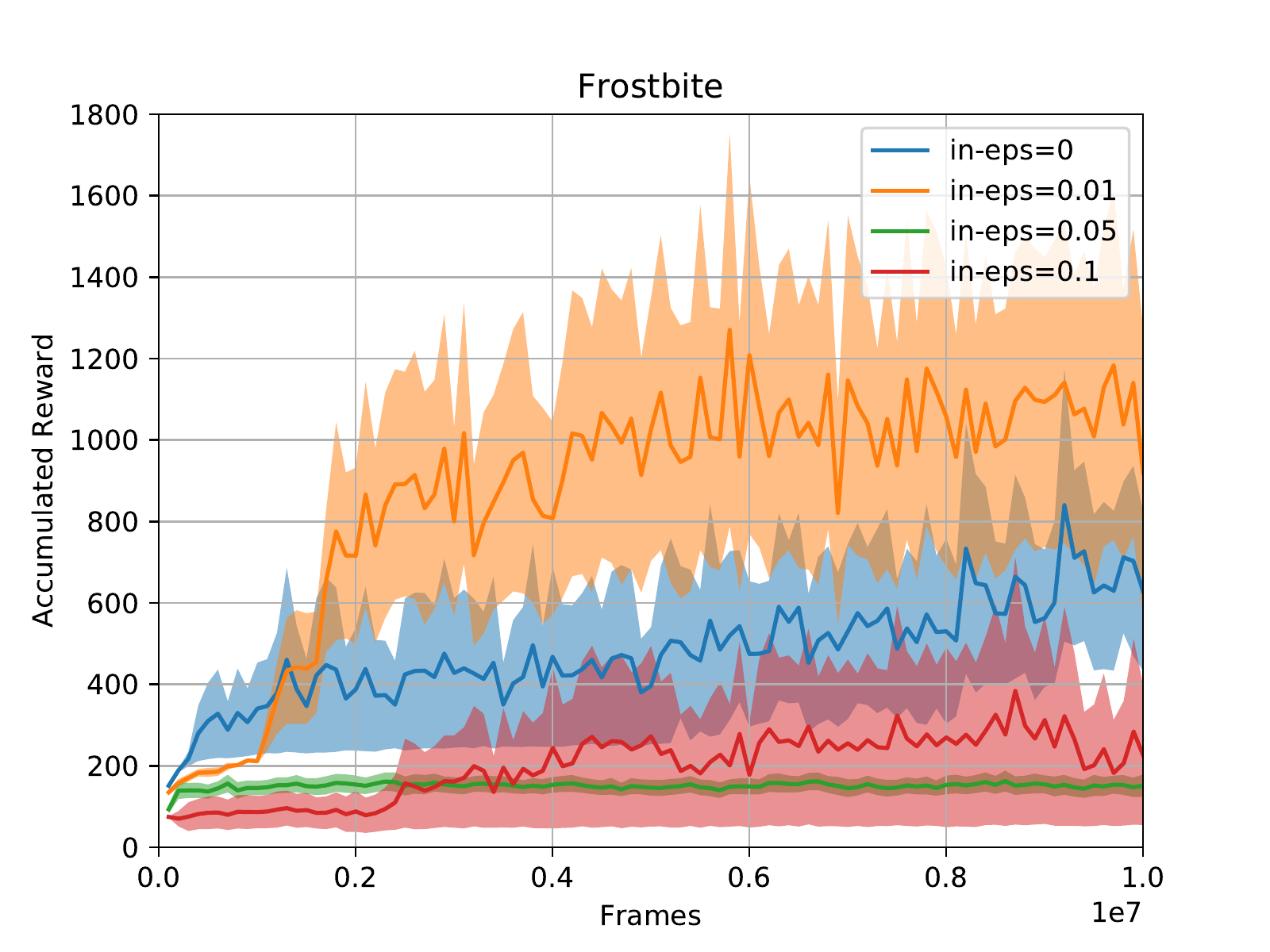}
        \label{fig:perf-frostbite}
    \end{subfigure}\\[-6.4ex]
    \begin{subfigure}{.49\textwidth}
        \centering
        \includegraphics[width=1\textwidth]{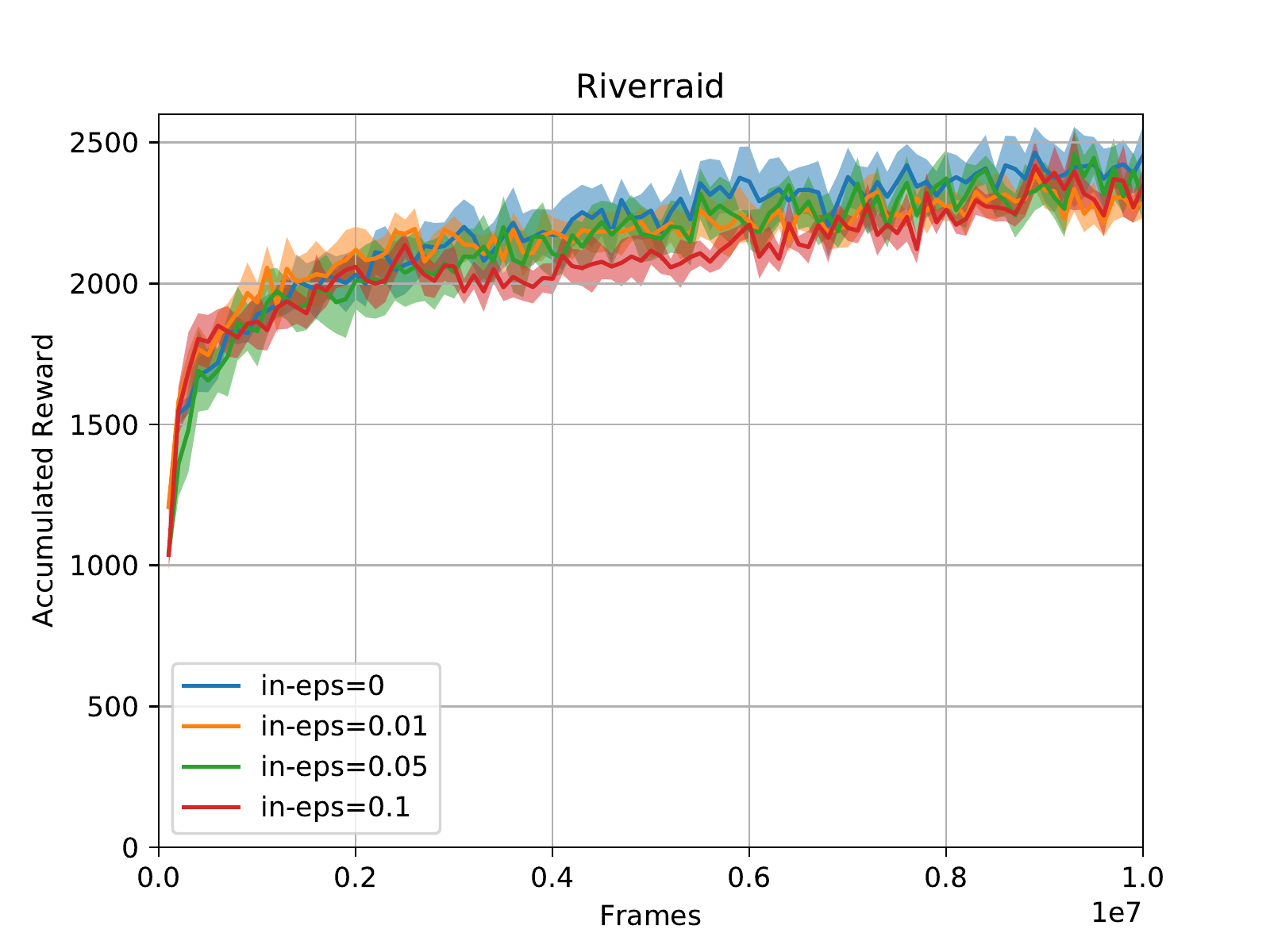}
        \label{fig:perf-riverraid}
    \end{subfigure}
    \begin{subfigure}{.49\textwidth}
        \centering
        \includegraphics[width=1\textwidth]{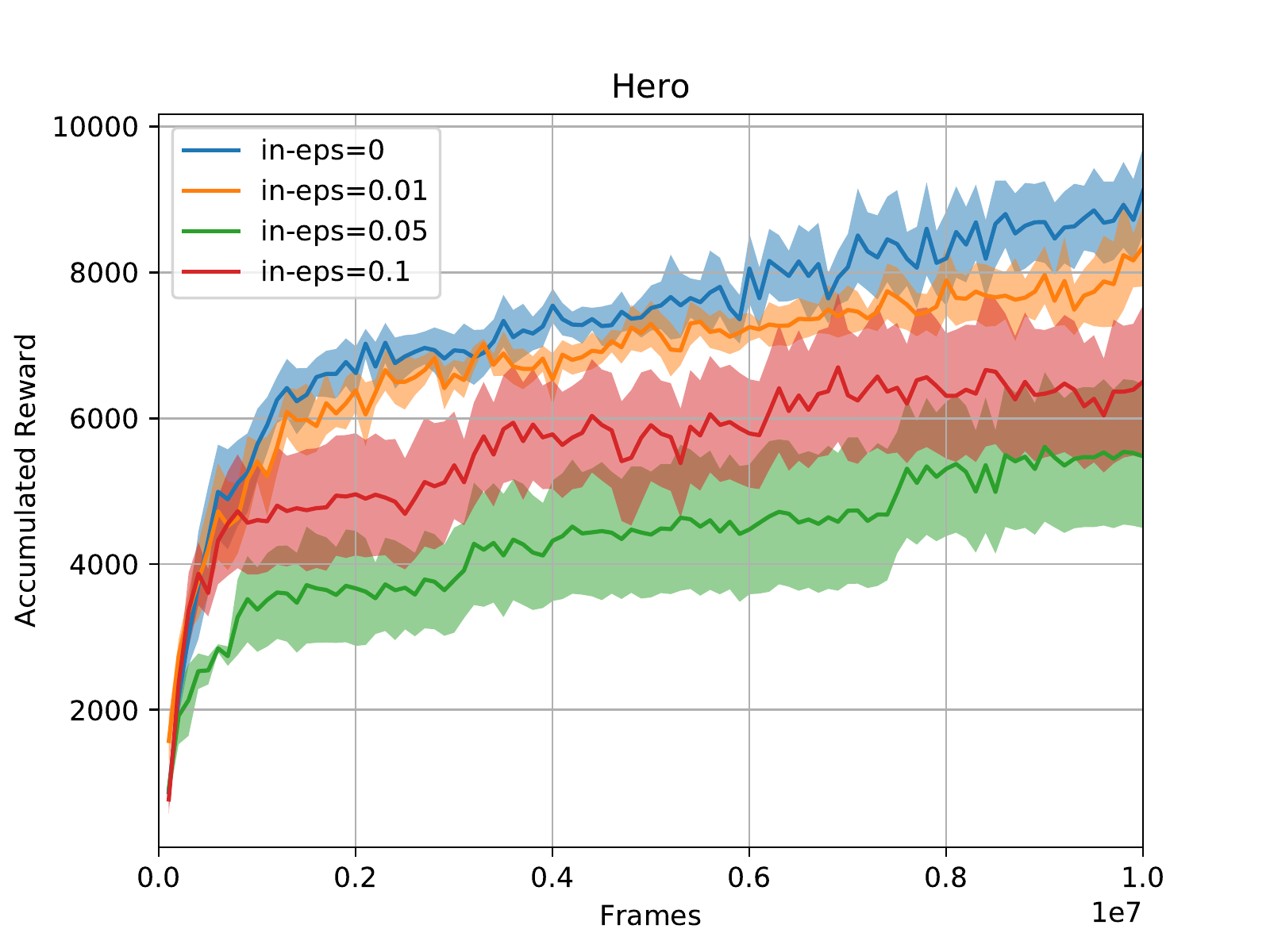}
        \label{fig:perf-hero}
    \end{subfigure}\vspace{-1.5\baselineskip}
    
    \caption{Average score on each game, considering ~100 thousand frames per epoch and a total of 10 million frames. Ms. Pacman, SpaceInvaders and River Raid have no significant impact on performance with different hyper-parameter settings.}
    \label{fig:perf}
\end{figure}

\begin{figure}[htb]
    \centering
    \begin{subfigure}{.49\textwidth}
        \centering
        \includegraphics[width=1\textwidth]{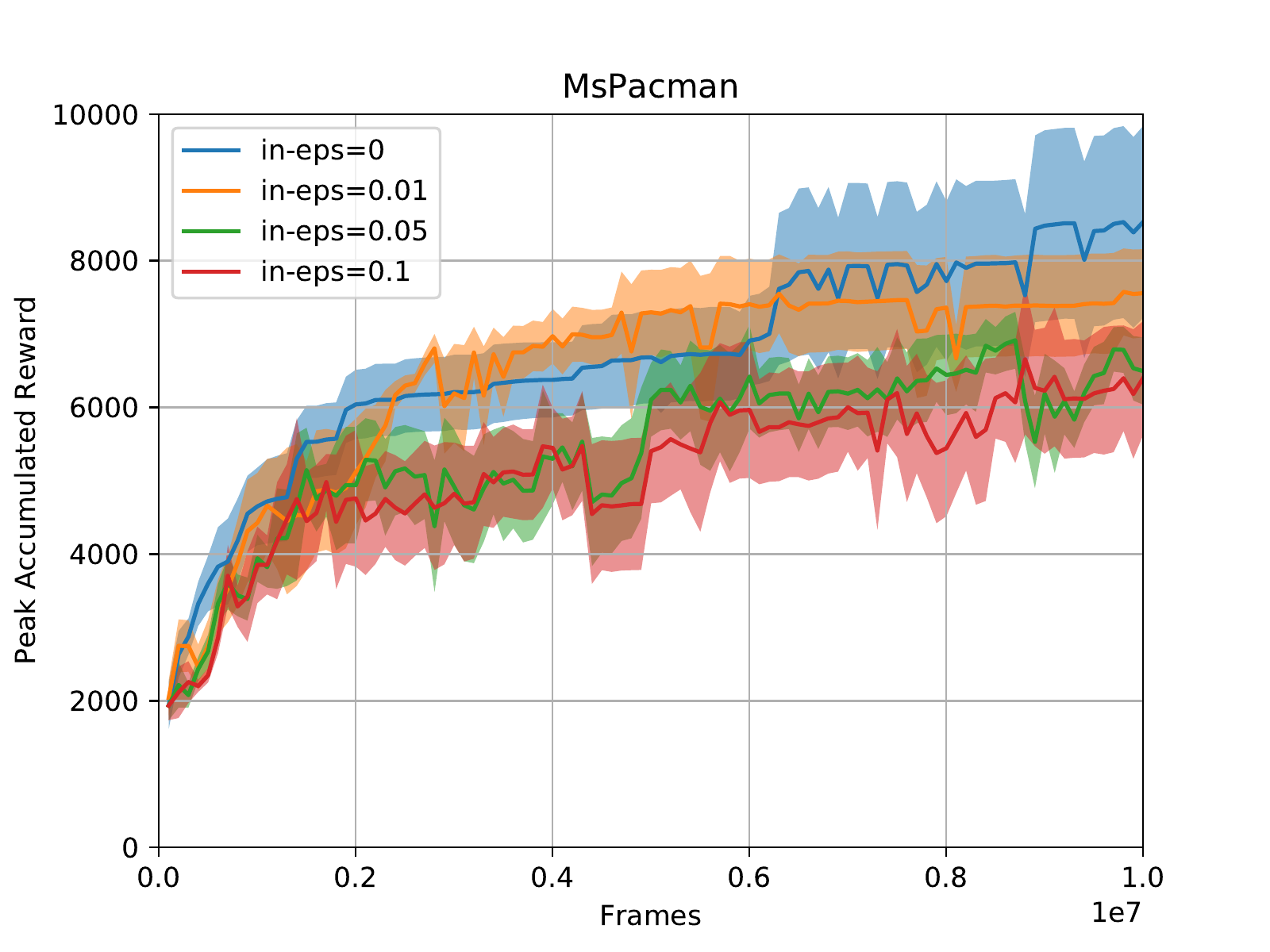}
        \label{fig:peak-mspacman}
    \end{subfigure}
    \begin{subfigure}{.49\textwidth}
        \centering
        \includegraphics[width=1\textwidth]{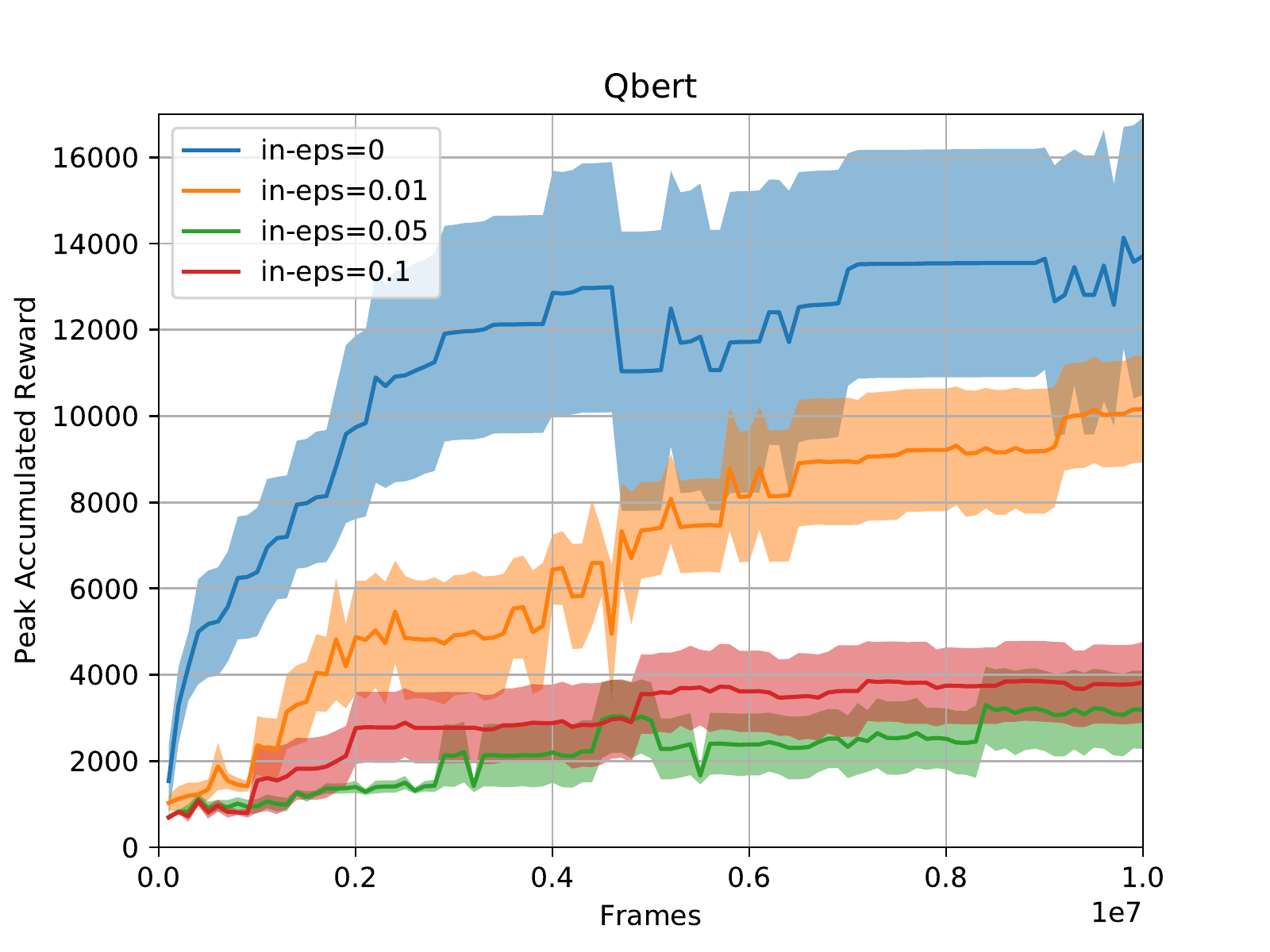}
        \label{fig:peak-qbert}
    \end{subfigure}\\[-6.4ex]
    \begin{subfigure}{.49\textwidth}
        \centering
        \includegraphics[width=1\textwidth]{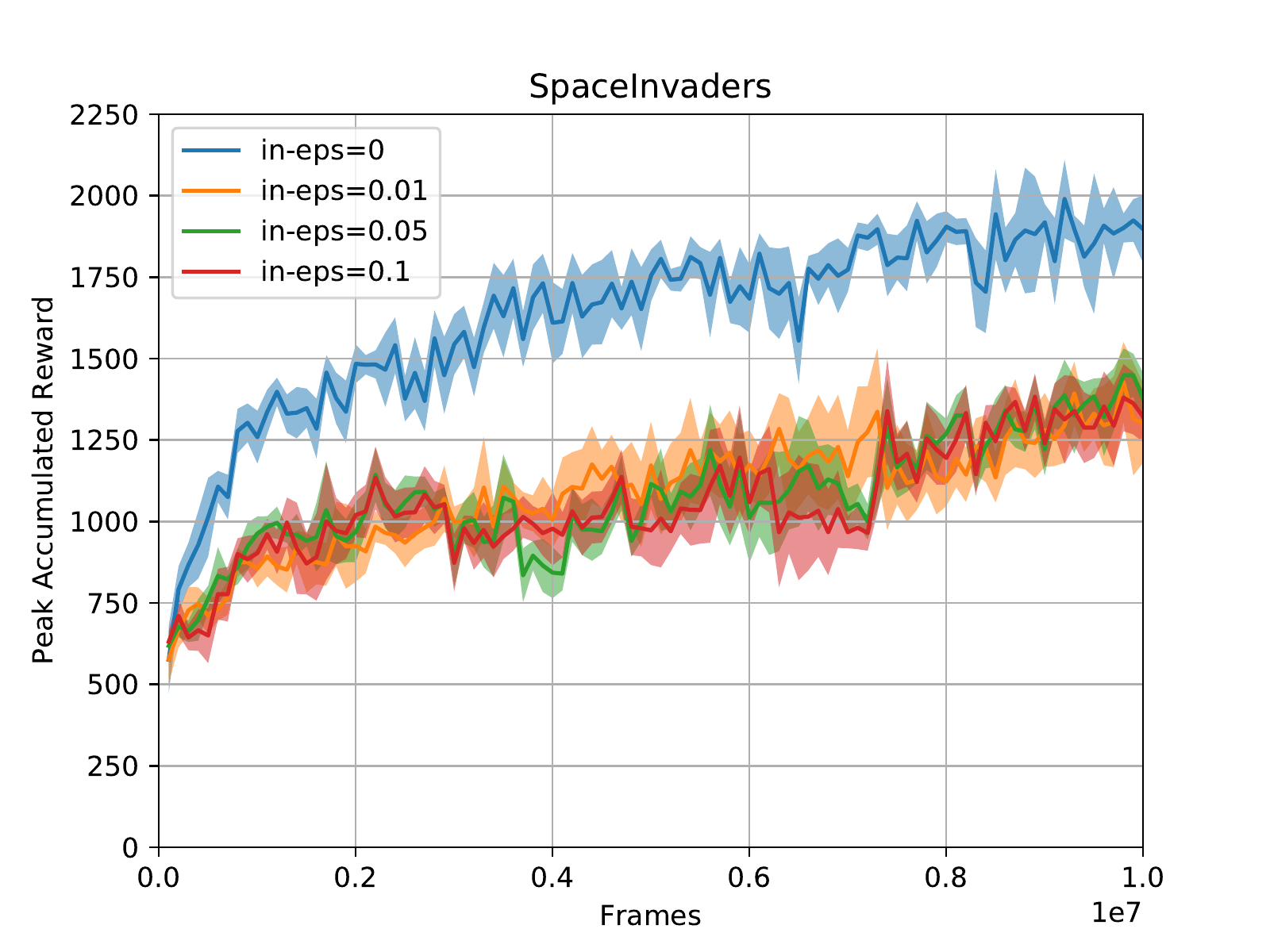}
        \label{fig:peak-spaceinvaders}
    \end{subfigure}
    \begin{subfigure}{.49\textwidth}
        \centering
        \includegraphics[width=1\textwidth]{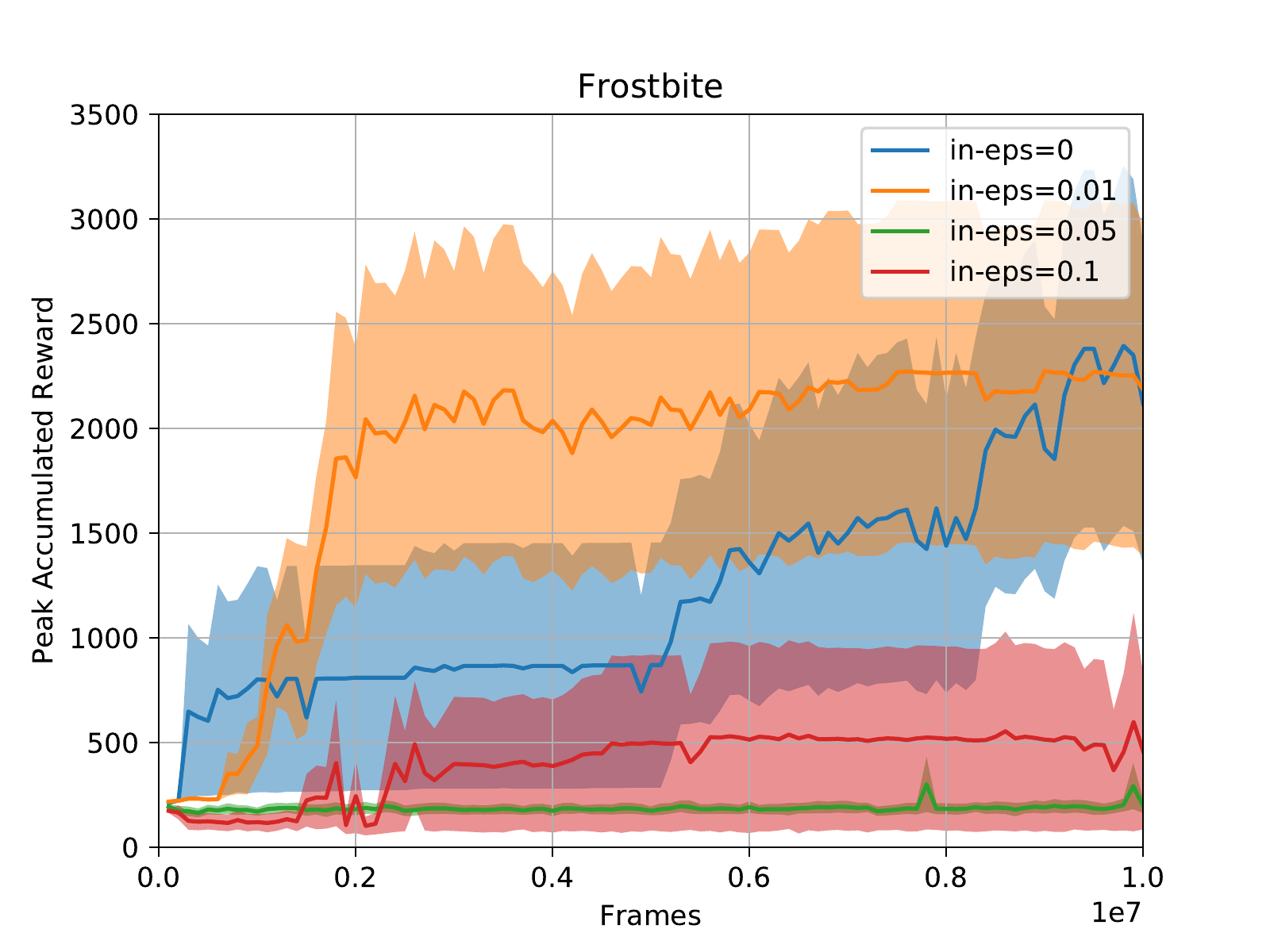}
        \label{fig:peak-frostbite}
    \end{subfigure}\\[-6.4ex]
    \begin{subfigure}{.49\textwidth}
        \centering
        \includegraphics[width=1\textwidth]{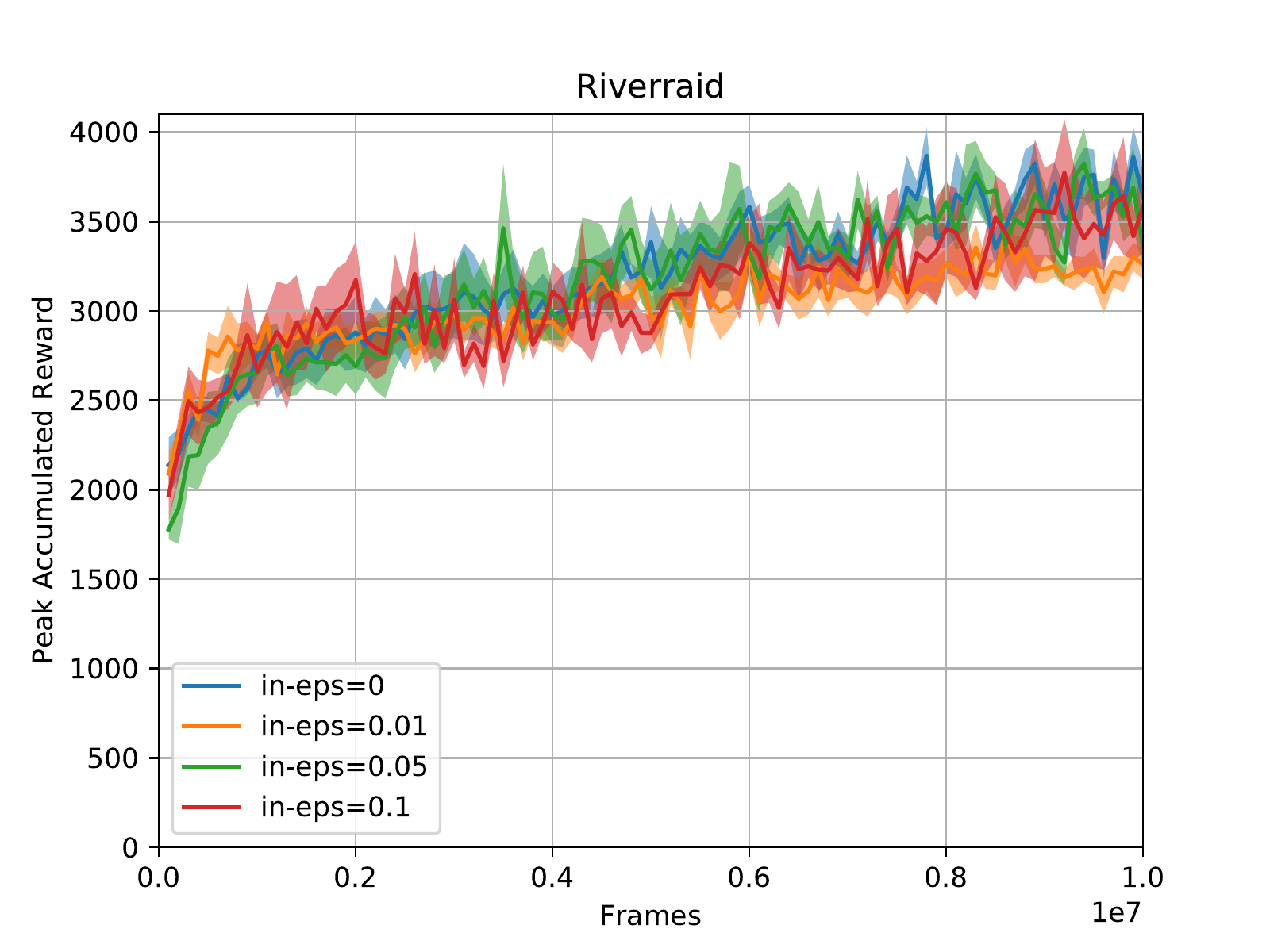}
        \label{fig:peak-riverraid}
    \end{subfigure}
    \begin{subfigure}{.49\textwidth}
        \centering
        \includegraphics[width=1\textwidth]{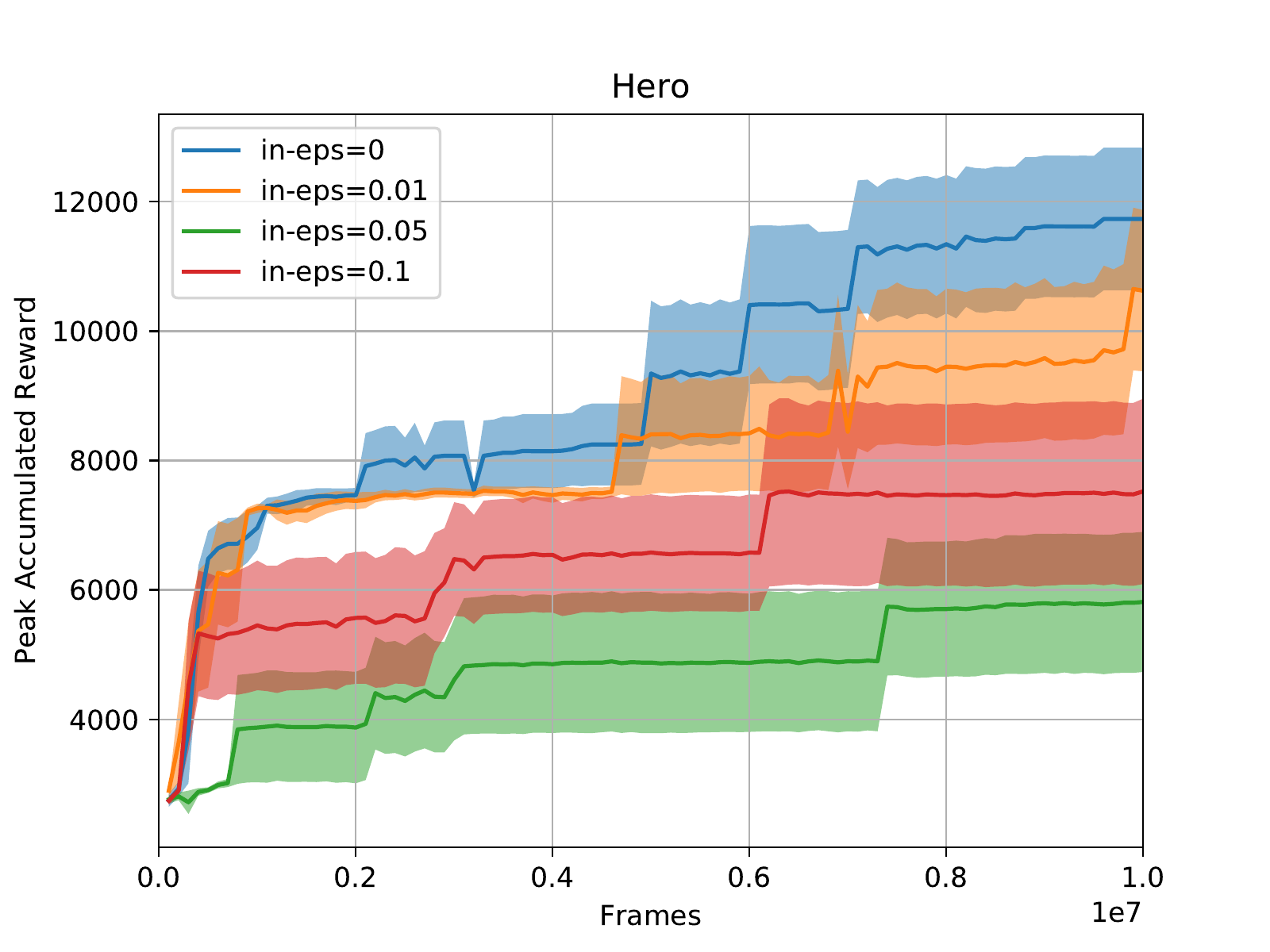}
        \label{fig:peak-hero}
    \end{subfigure}\vspace{-1.5\baselineskip}
    
    \caption{Peak score on each game, considering ~100 thousand frames per epoch and a total of 10 million frames. Only River Raid is insensitive to the different hyper-parameter settings.}
    \label{fig:peak}
\end{figure}

\subsection{Results}

\begin{figure}[htb]
    \centering
    \begin{subfigure}{.49\textwidth}
        \centering
        \includegraphics[width=1\textwidth]{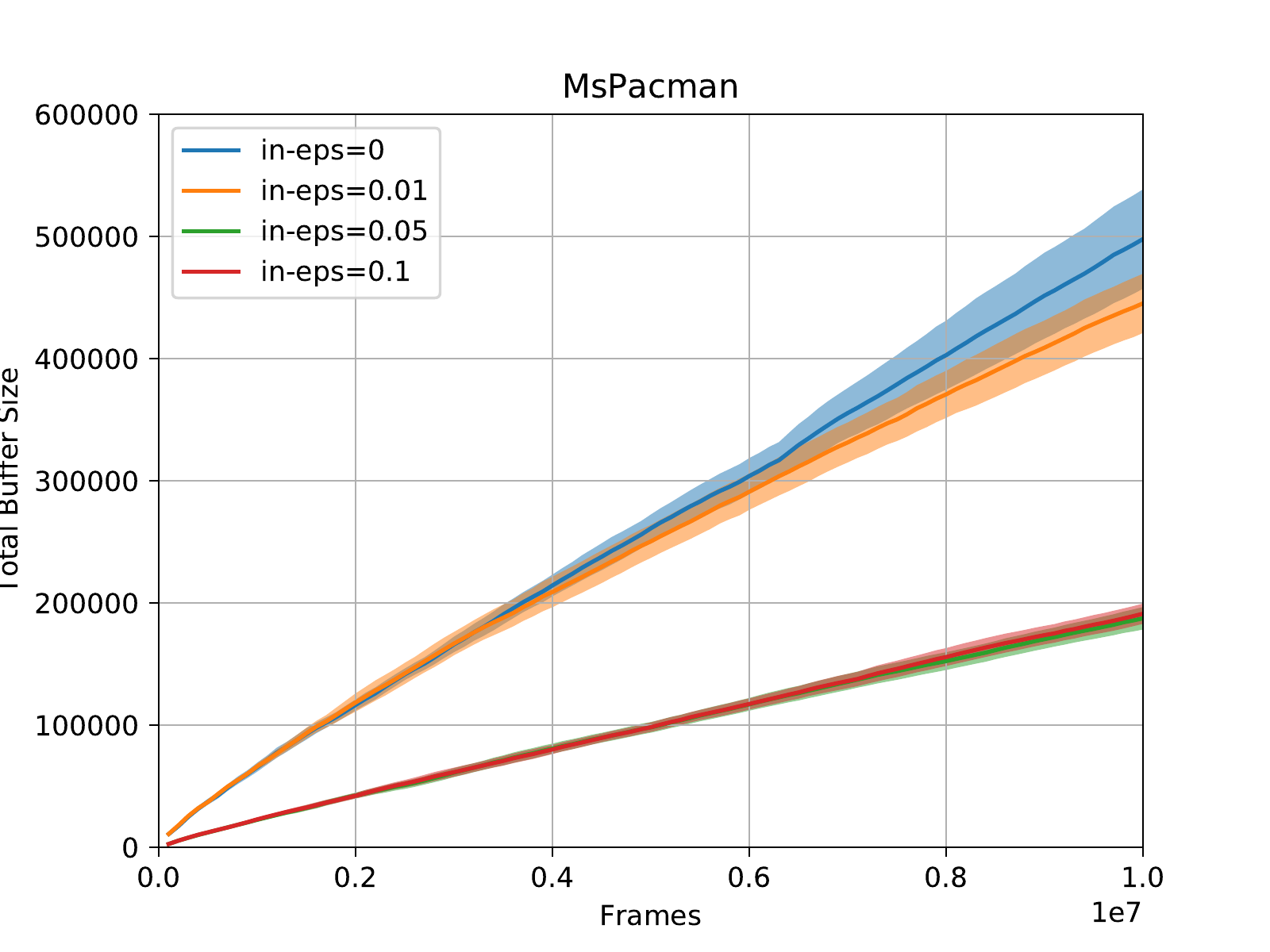}
        \label{fig:size-mspacman}
    \end{subfigure}
    \begin{subfigure}{.49\textwidth}
        \centering
        \includegraphics[width=1\textwidth]{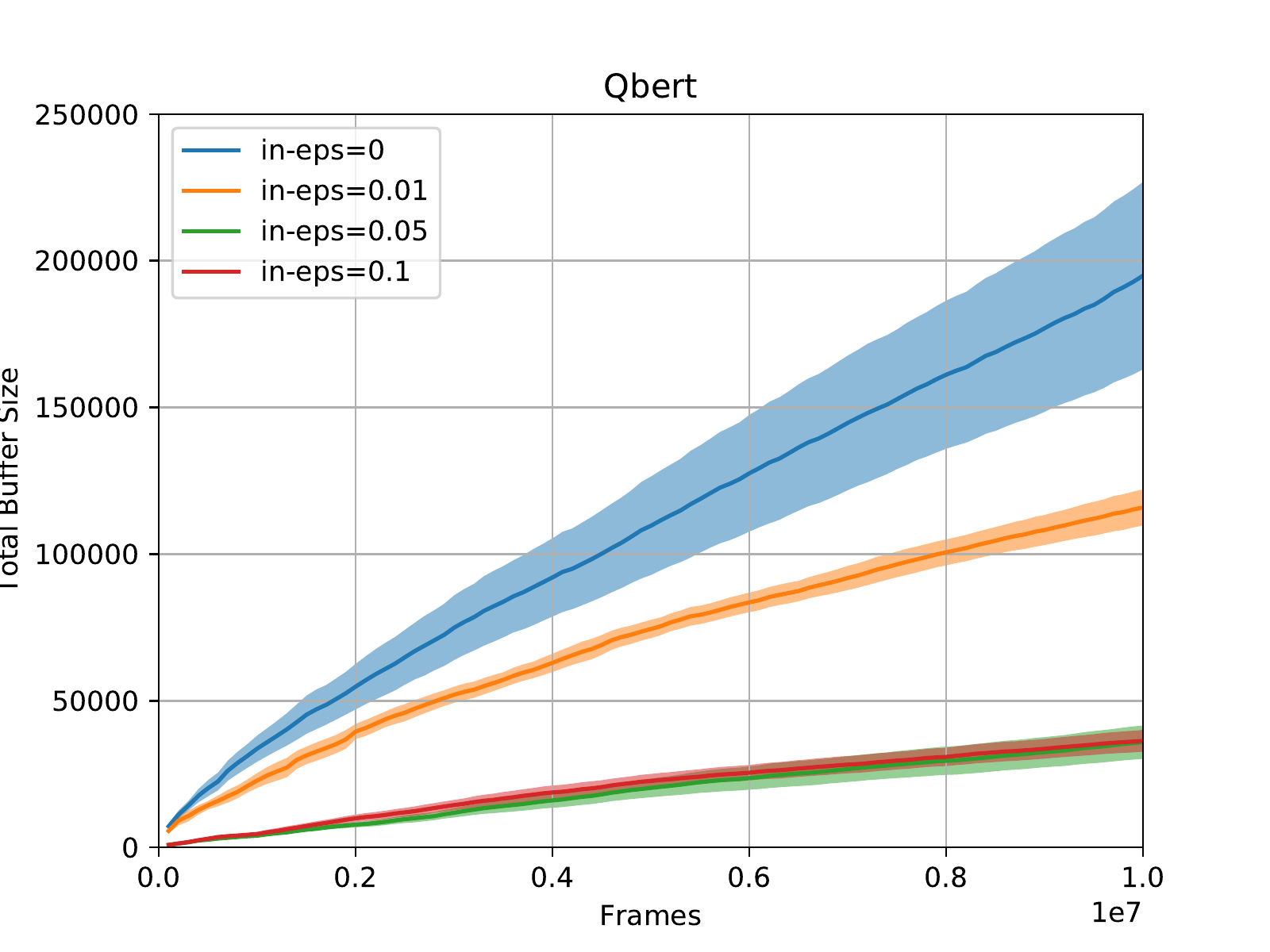}
        \label{fig:size-qbert}
    \end{subfigure}\\[-6.4ex]
    \begin{subfigure}{.49\textwidth}
        \centering
        \includegraphics[width=1\textwidth]{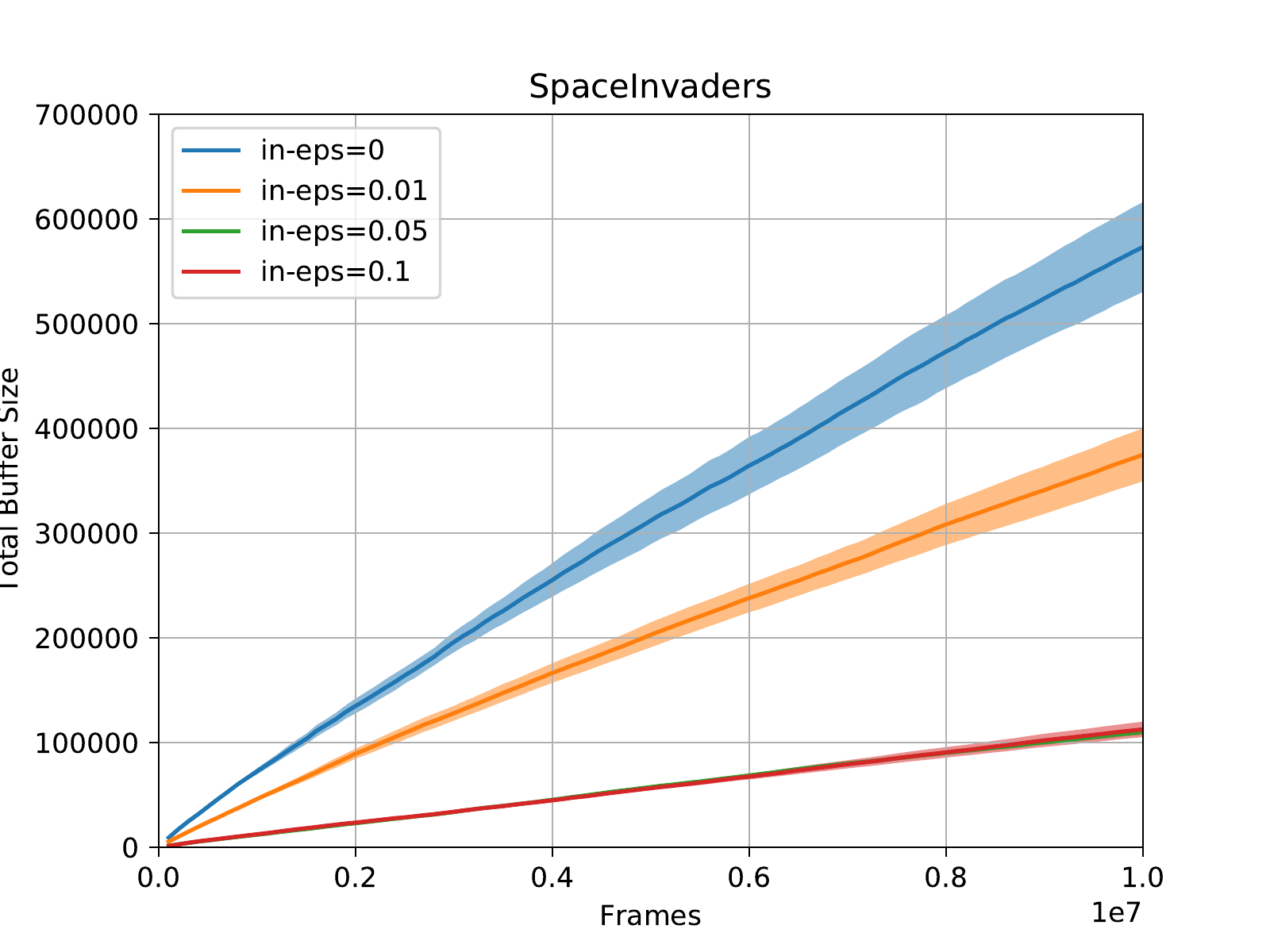}
        \label{fig:size-spaceinvaders}
    \end{subfigure}
    \begin{subfigure}{.49\textwidth}
        \centering
        \includegraphics[width=1\textwidth]{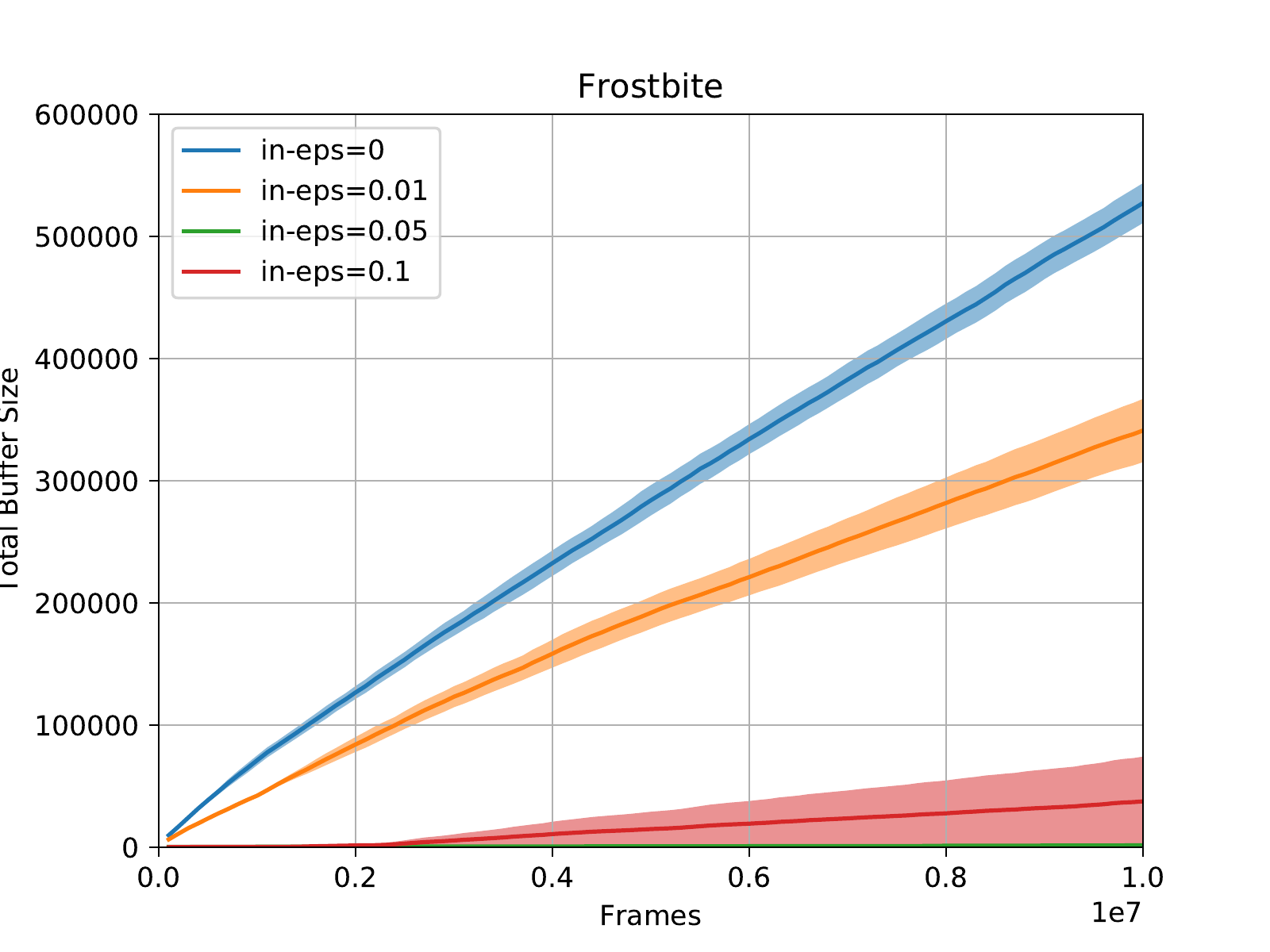}
        \label{fig:size-frostbite}
    \end{subfigure}\\[-6.4ex]
    \begin{subfigure}{.49\textwidth}
        \centering
        \includegraphics[width=1\textwidth]{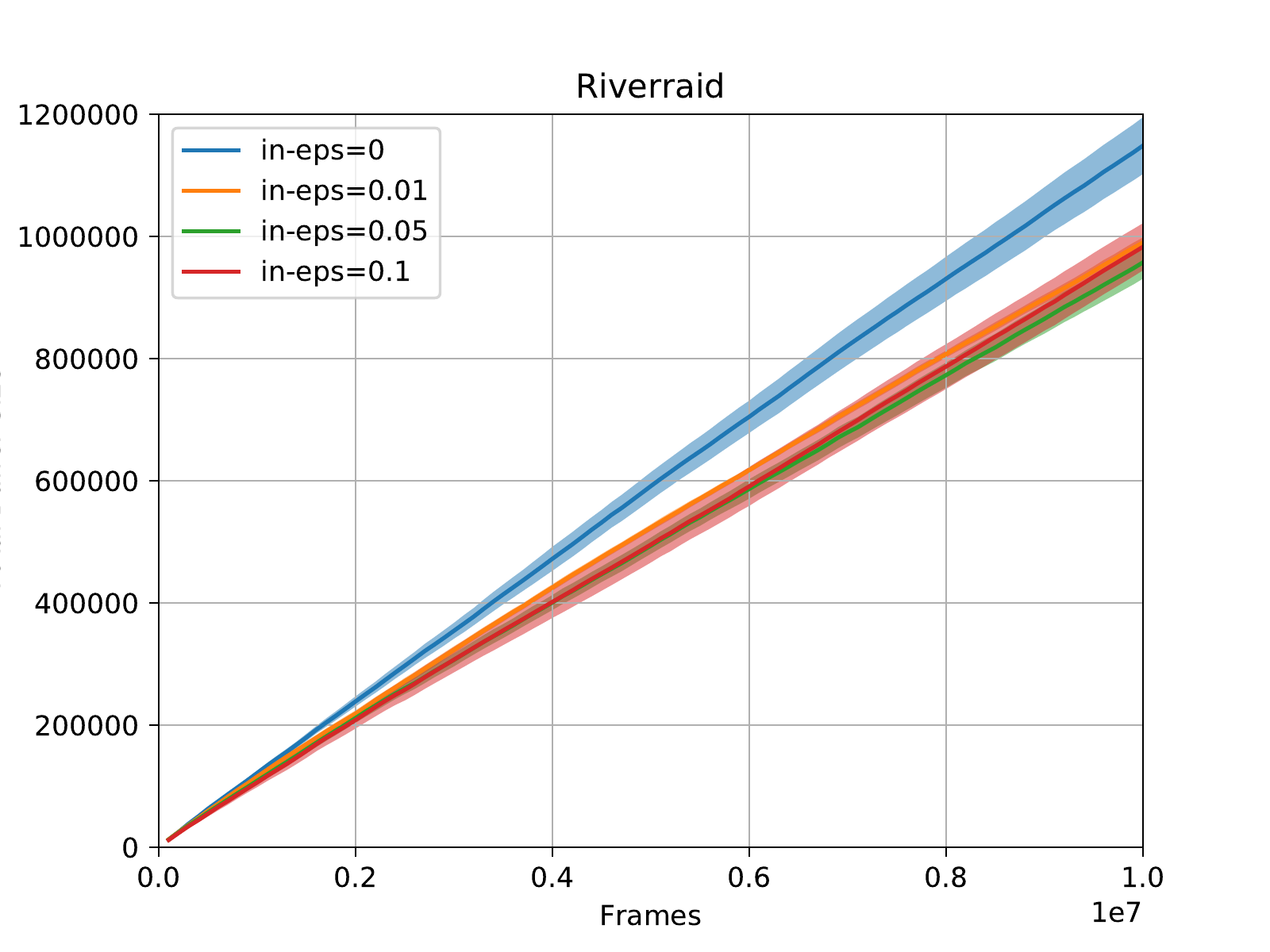}
        \label{fig:size-riverraid}
    \end{subfigure}
    \begin{subfigure}{.49\textwidth}
        \centering
        \includegraphics[width=1\textwidth]{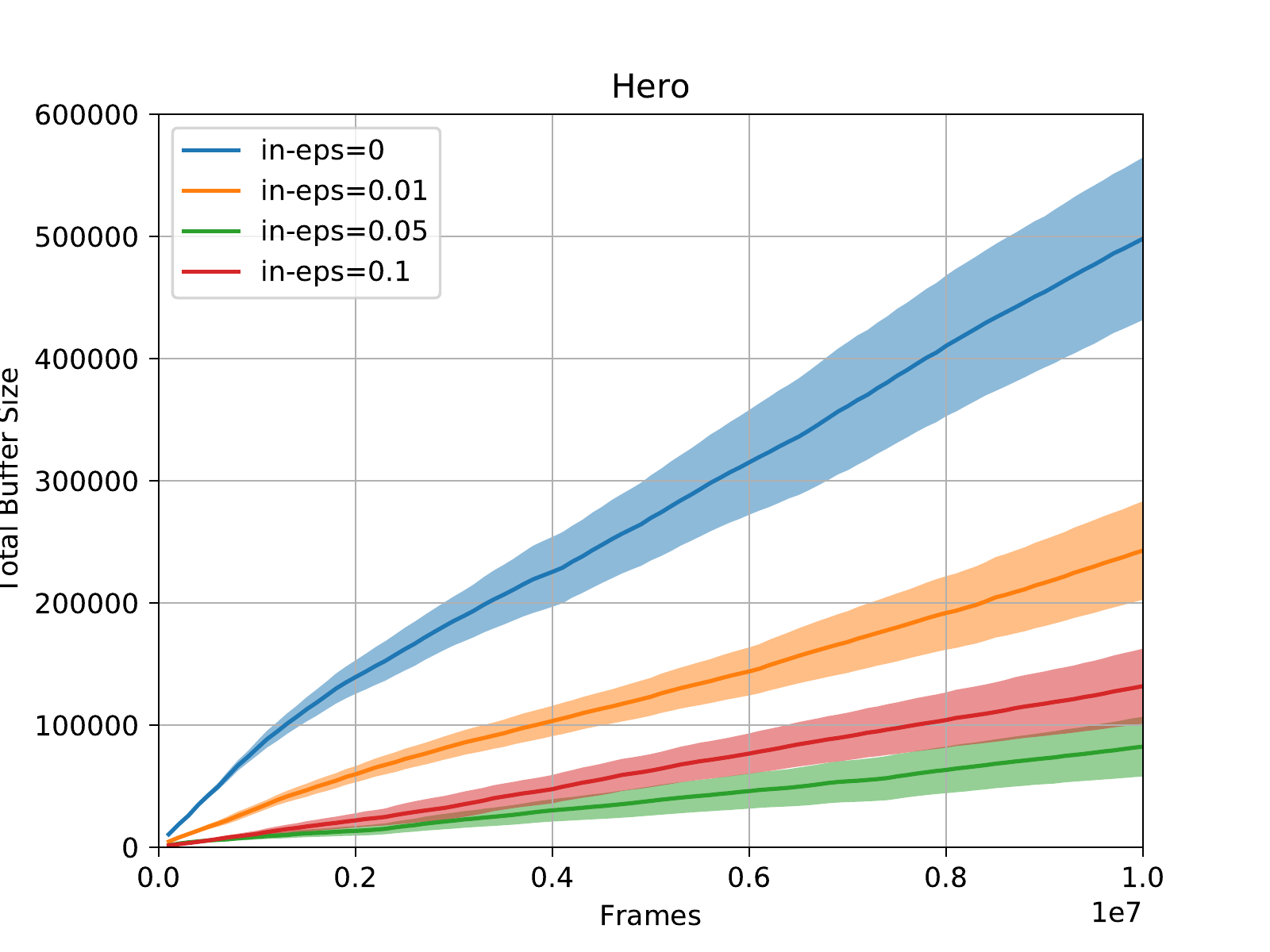}
        \label{fig:size-hero}
    \end{subfigure}\vspace{-1.5\baselineskip}
    
    \caption{Total buffer size (i.e., the sum of all action buffers) on each game, considering ~100 thousand frames per epoch and a total of 10 million frames. All settings have an impact on buffer sizes. Dramatic reductions in buffer size generally imply loss of performance, as can be observed in Q*bert, Frostbite and Hero, according to figures \ref{fig:perf} and \ref{fig:peak}.}
    \label{fig:size}
\end{figure}

Figures \ref{fig:perf} and \ref{fig:peak} show mean curves and standard error intervals for average and peak performance in the games, respectively. Figure \ref{fig:size} shows mean curves and standard error intervals for total buffer size, i.e., the sum of all action buffers for each game. It can be observed that, albeit not always giving great improvements in terms of size, the $\varepsilon_{in} = 0.01$ setting is a quite safe option regarding average performance, even giving better results in Frostbite. In the case of Ms. Pacman, Space Invaders, and River Raid, no setting incurred a significant loss of average performance. The same behavior can not be observed in the peak performance case for Q*bert and Space Invaders, as any level of state aggregation incurs in performance loss. It seems that fine-grained state information is necessary to achieve very high scores in these games, even if only occasionally (projectile size in Space Invaders being evidence for this, as well as small brightness differences in Q*bert blocks when discarding color information, as is the case). Even vanilla MFEC produces peak performance much above its average, suggesting that it struggles to return to past high rewarding states after discovering them (this problem is approached by \cite{ecoffet2020first}, which could signal promising venues of improvement for episodic control as well). River Raid shows its peculiar behavior in all metrics: there is no significant loss of average or even peak performance for all settings, but the size reduction is not dramatic in any case. We hypothesize that this is due to the continuous vertical scrolling in the game, which results in every state being completely different in pixel space (note that this game produces the largest buffers among the experiments). In any case, it is clear that the hyper-parameters are problem-dependent and should be tuned for each task. A possible solution, not explored in this work, would be to automatically find good values for $\varepsilon_{in}$ and $\varepsilon_{out}$ in a data-driven manner, by observing the distribution of states.

\section{Conclusions}
\label{sec:conclusions}

In previous sections, a simple heuristic for reducing the number of stored states in MFEC was presented. While MFEC stores all unique states encountered during learning, we proposed to apply thresholds for storing states according to a maximal dissimilarity criterion. These dissimilarity thresholds should apply to both input (state) and output (Q-value) spaces to avoid aliasing (multiple optimal actions for single stored states). States too similar, with close enough Q-values are aggregated, reducing memory and computational requirements. Besides requiring less memory, it allows for faster k-nearest neighbor lookups, as the number of candidates is reduced.

Experiments were performed on six Atari games, showing that conservative choices of thresholds (aggregating only very similar states) successfully reduce computational demands, while resulting in no significant loss of performance. It could be observed that different threshold values have different effects on each game. This is related to how each state space has its own distribution characteristics. A method for automatic tuning could make the heuristic more robust and general. We suggest that maintaining variance estimates for each aggregated state could be a simple way to define a per-state similarity threshold that would adapt to each task. Another possibility would be to avoid state aggregations that result in policy changes (a change in $argmax_a Q$). 

Some interesting directions for future research can be conjectured from this study. A possible simple modification to the algorithm would be to introduce distance-weighted approximations and remove the hard $k$ limit of neighbors (our preliminary experiments in this direction were not promising), or even combine both criteria. This, in turn, would allow the contributions of this work to be applied to NEC as well. Introducing visit counters into each state should be trivial for episodic learning in general and would allow for efficient count-based exploration \cite{bellemare2016unifying}. Also regarding efficient exploration, we observed the $k$ hyper-parameter to have a high impact on the quality of exploration, being responsible for the low reliance on the $\varepsilon$ exploration rate hyper-parameter (which can be set to very low values compared to other reinforcement learning methods). This suggests that the role of $k$ could be beyond a mere approximation of Q-values. In fact, newly explored regions contain a low number of states, forcing kNN to include distant states into its estimate, producing a form of implicit exploration in novel states and exploitation in well-explored regions. This could explain why distance weighting produced no satisfactory results in some of our tentative experiments, but more in-depth studies are required since other works suggest distance weighting to be beneficial \cite{agostinelli2019memory}. 

\section*{Acknowledgments}

We gratefully acknowledge the support of NVIDIA Corporation with the donation of the Titan Xp GPU used for this research.

\bibliographystyle{unsrt}  
\bibliography{references}  

\end{document}